\documentclass[10pt,twocolumn,letterpaper]{article}

\usepackage[review]{arxiv}     

\usepackage{url}
\usepackage{graphicx}
\usepackage{booktabs}
\usepackage{siunitx}
\usepackage{multirow}
\usepackage{array}
\usepackage[symbol]{footmisc}

\newcommand{\figref}[1]{Figure~\ref{#1}}

\definecolor{cvprblue}{rgb}{0.21,0.49,0.74}
\usepackage[pagebackref,breaklinks,colorlinks,allcolors=cvprblue]{hyperref}

\def\paperID{*****} 
\def\confName{CVPR}
\def\confYear{2025}

\title{Human Gaze Boosts Object-Centered Representation Learning}

\author{\textbf{Timothy Schaumlöffel}\text{*}\\
Goethe University Frankfurt\\
The Hessian Center AI\\
{\tt\small schaumloeffel@em.uni-frankfurt.de}
\and
\textbf{Arthur Aubret}\text{*}\\
Frankfurt Institute for Advanced Studies\\
Xidian-FIAS international Joint Research Center\\
{\tt\small aubret@fias.uni-frankfurt.de}
\and 
\textbf{Gemma Roig}\text{$\dagger$}\\
Goethe University Frankfurt\\
The Hessian Center AI\\
{\tt\small roig@cs.uni-frankfurt.de}
\and 
\textbf{Jochen Triesch}\text{$\dagger$}\\
Frankfurt Institute for Advanced Studies\\
Goethe University Frankfurt\\
{\tt\small triesch@fias.uni-frankfurt.de}
}

\begin{document}
\maketitle
\begin{abstract}
Recent self-supervised learning (SSL) models trained on human-like egocentric visual inputs substantially underperform on image recognition tasks compared to humans. These models train on raw, uniform visual inputs collected from head-mounted cameras. This is different from humans, as the anatomical structure of the retina and visual cortex relatively amplifies the central visual information, \textit{i.e.} around humans’ gaze location. This selective amplification in humans likely aids in forming object-centered visual representations. Here, we investigate whether focusing on central visual information boosts egocentric visual object learning. We simulate 5-months of egocentric visual experience using the large-scale Ego4D dataset and generate gaze locations with a human gaze prediction model. To account for the importance of central vision in humans, we crop the visual area around the gaze location. Finally, we train a time-based SSL model on these modified inputs. Our experiments demonstrate that focusing on central vision leads to better object-centered representations. Our analysis shows that the SSL model leverages the temporal dynamics of the gaze movements to build stronger visual representations. Overall, our work marks a significant step toward bio-inspired learning of visual representations.
\end{abstract}

\section{Introduction}

Current cutting-edge vision models outperform humans by a large margin on object recognition \cite{russakovsky2015imagenet} and begin to reach humans' level on out-of-distribution object recognition \cite{geirhos2021partial}. However, it does not mean that vision models are better at learning than humans, as these vision models train on far more diverse images and labels than humans. For instance, the (relatively small) widespread ImageNet-1k dataset includes tens of thousands of dogs, as well as thousands of images from uncommon categories like sea snakes, etc\dots To be compared with humans, vision models should train on the same egocentric visual inputs experienced by humans without abundant supervision. One way to do that is to train self-supervised learning (SSL) models on videos collected from head-mounted cameras \cite{orhan2024learning}. However, these egocentric SSL models still lag behind humans by $[20,50]\%$ in recognition accuracy depending on the object-centered dataset \cite{orhan2023scaling,orhan2024learning}.


In practice, egocentric data largely differs from images in objects-centered datasets. Egocentric videos mostly show images of scenes, like a living room filled with tens of objects whereas object-centered datasets often show one or two big objects. This is important as pre-training on images of scenes transfers poorly on object-centered datasets \cite{grill2020bootstrap} (Appendix~F). This problem may be even more exacerbated with egocentric data, as egocentric SSL models tend to learn representations that are less object-centered and more background-sensitive \cite{orhan2024learning,aubret2022toddler}. Overall, this suggests that raw videos extracted from head-mounted cameras are inadequate for learning strong object-centered representations.

The stimuli received by humans' visual cortex structurally differ from such egocentric videos. The anatomy of the retina relatively amplifies the information located in the center of the field of view \cite{anstis1974chart,wassle1989cortical}, meaning that high and intermediate acuity processing occurs only within several degrees from the center of the visual field, \textit{i.e.} in central vision. As a consequence, central vision plays a crucial role in the formation of visual representations in areas of the visual cortex related to semantic information \cite{quaia2024object, yu2015representation}. To compensate for the relatively low acuity in peripheral vision, humans need to actively move their gaze onto different objects to parse their environment. Since human gazes are naturally attracted by salient objects, the visual sequence in central vision may most of the time present a few centered and big objects, similar to object-centered datasets. 

\begin{figure*}
    \centering
    \includegraphics[width=1.0\linewidth]{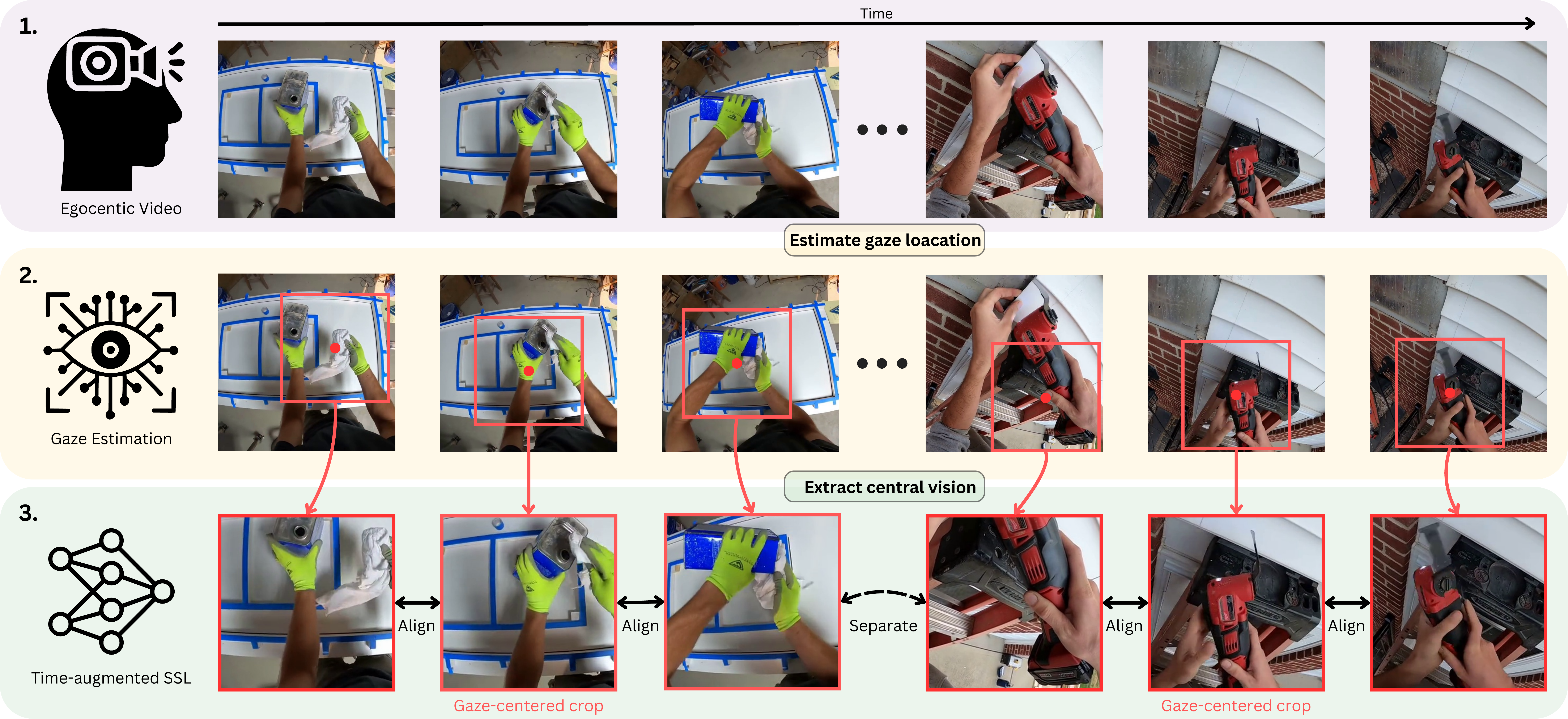}
    \caption{Illustration of our data generation and model training approach. (1) We extract frames from the egocentric dataset Ego4D \cite{grauman2022ego4dworld3000hours}. (2) For each frame, we predict the human gaze location (red dot) using a state-of-the-art model \cite{Lai_2022_BMVC}. (3) We train a time-augmented SSL model to align representations of gaze-centered crops (red rectangle) extracted from close-in-time frames.}
    \label{fig:main}
\end{figure*}

In this paper, we investigate whether focusing on the visual information located in humans' gaze locations can boost object-centered representations in egocentric SSL. We use Ego4D \cite{grauman2022ego4d}, a dataset that contains 3,670 hours of videos collected with head-mounted cameras. This dataset contains gaze locations on only a subset of videos (45 hours). Thus, we apply a state-of-the-art model of human gaze prediction to generate gaze locations on the rest of the dataset \cite{Lai_2022_BMVC}. To simulate the importance of central vision in humans, we propose simply cropping a subpart of the image centered on the location of the gaze. Finally, we train a variant of an SSL model, which trains visual representations to slowly change, on the resulting visual sequence for one single epoch.

Our experiments demonstrate that pre-training an SSL model on central vision boosts category, fine-grained and instance object recognition with linear probing, compared to training with the whole field of view. Our analysis suggests that this boosts roots in learned representations relying more on foreground object information than background information. In addition, we find that learning temporal slowness is critical to the learning process and that gaze movements specifically support visual learning. Overall, our work exhibits the importance of using central vision to learn visual representations from an egocentric visual experience. Thus, we make an important step towards learning strong visual representations using similar data as humans.



\section{Related works}

\paragraph{Egocentric SSL.} The increased availability of datasets collected with head-mounted cameras \cite{grauman2022ego4d,sullivan2021saycam,long2024babyview,greene2024visual,ma2024nymeria} recently induced a surge for training egocentric SSL \cite{orhan2024self}. To the best of our knowledge, all these approaches train egocentric SSL models using the entire field of view captured by head-mounted cameras. For instance, \cite{orhan2024self} trained a MAE \cite{he2022masked}, MUGS \cite{zhou2022mugs} and DINO \cite{caron2021emerging} on egocentric data and found that training on 10\% of ImageNet was better than training on 200 hours of egocentric data. Prior work also found that endowing SSL models with representations that slowly change over time can slightly boost the learning process \cite{orhan2020self}. Similar egocentric SSL models were recently trained on a larger scale on the full Ego4D dataset \cite{long2024babyview} and a combination of datasets \cite{emin2024hvm}. A similar line of work leverages egocentric data to train vision models useful for solving robotic tasks. Among these models, VC-1 is an MAE \cite{majumdar2023we} trained on combinations of egocentric and third-person videos. R3M \cite{nairr3m} and VIP \cite{mavip} both learn slowly changing representations on Ego4D and R3M additionally aligns visual representations with language utterances. We show in \cref{sec:expe} that training on gaze-based central vision elicits better object representations. Other works try to extract the correspondences between objects' views in videos to learn visual representations \cite{jabri2020space,venkataramanan2024imagenet,salehi2023time,parthasarathy2023self,gordon2020watching}. Here, we are rather interested in understanding how the biological importance of central vision may impact egocentric SSL. 

\paragraph{Time-based SSL.} Many works previously proposed to learn similar representations for close-in-time visual inputs \cite{wiskott2002slow,foldiak1991learning}. More recently, this learning principle has been integrated into mainstream SSL methods \cite{aubret2022time}. However, these works do not leverage in-the-wild egocentric data. A first line of works mimics humans' visual experience with synthetic \cite{aubret2022time,schaumloffel2023caregiver} or curated \cite{aubret2024self,sanyal2023computational,aubret2024learning} visual sequences of interactions with objects. A second line of works uses different kinds of videos, like third-person ones \cite{sermanet2018time}, videos recorded by a car \cite{jayaraman2015learning,jayaraman2016slow}, movie video clips \cite{jayaraman2016slow}, chicks egocentric perspective \cite{pandey2024vision} or object-tracking datasets \cite{xu2021rethinking}. In this paper, we study the learning of time-augmented SSL when trained on central vision during the daily life of humans.



\section{Method}

We aim to study the potential impact of focusing on central vision when training with egocentric SSL. We use the largest-to-date dataset of egocentric videos (Ego4D) and generate gaze locations with a state-of-the-art model of human gaze prediction. To simulate the biological importance of central vision, we simply crop the visual area around a gaze location (\Cref{sec:dataset}). This creates a sequence of visual inputs that feed a time-augmented variant of an SSL method, which is described in \Cref{sec:model}. \figureautorefname~\ref{fig:main} illustrates the main steps of the pipeline. Finally, we explain how we evaluate our representation with respect to different semantic aspects of objects in \Cref{sec:evaluation}.

\subsection{Dataset} \label{sec:dataset}

\paragraph{Human-like egocentric visual experience.} To simulate the visual experience of humans, we use the
Ego4D dataset \cite{grauman2022ego4dworld3000hours}. This dataset contains 3,600 hours of videos collected through head-mounted cameras, which corresponds to approximately 5 months of visual experience. 931 participants coming from 74 worldwide locations wore a camera for one to ten hours. Thus, Ego4D arguably represents much more than 5 months of experience for a single average human in terms of diversity, although it is hard to make precise estimates. We use videos with a resolution of $540\times540$ pixels and extract their frames at approximately 5 fps, following previous findings that a higher fps does not boost the learning process \cite{sheybani2024curriculum}. 

\paragraph{Gaze location.} During frame extraction, we create small clips of 5 seconds (25 frames) that we sequentially load into memory. We gather 24 frames of these 25 frames and split them into three sequences of 8 frames. For Ego4D videos recorded with an eye-tracker (45 hours), we do not further process the frames and associate them with ground-truth gaze location. For all other videos, we feed each sequence into GLC, a state-of-the-art model of human gaze prediction trained on the Ego4D subset that contains gaze locations \cite{Lai_2022_BMVC}. This model uses spatio-temporal information to generate a saliency map for each of the 8 frames. Compared to single-image saliency models \cite{riche2016bottom}, this allows the model to generate a temporally consistent gaze location and to leverage more cues (e.g. motion). For each frame, we take as gaze location the position of the most salient pixel $(x_g, y_g)$. Our final preprocessed dataset contains 64,380,024 images.

\paragraph{Bio-inspired focus on central vision.}
To simulate the importance of central vision in humans, we propose to crop a $N \times N$ squared area centered on the gaze location for each frame. The crop boundaries may go beyond the image boundaries; in this case, we minimally shift the crop such that its boundaries remain in the image. Precisely, we compute the corrected gaze location $(x_g^{\texttt{cor}},y_g^{\texttt{cor}})$ as
\begin{align}
    x_g^{\texttt{cor}} = x_g - \max(0,x_g) + \frac{N}{2}- 540) - \min(0, x_g - \frac{N}{2})\nonumber,\\
    y_g^{\texttt{cor}} = y_g - \max(0,y_g) + \frac{N}{2} - 540) - \min(0, y_g - \frac{N}{2})\nonumber,
\end{align}
where $540$ denotes the resolution of our frames. We study how $N$ impacts the learning process in \Cref{sec:expe}.


\subsection{Learning model}\label{sec:model}
Our long-term objective is to learn good representations when using similar data as humans. Since most human visual experience is unsupervised, we employ SSL models. These models learn high-level visual representations without any explicit supervision, like human-provided labels. 
In this work, we focus specifically on the third version of Momentum Contrast (MoCoV3) \cite{chen2021empirical}, which is one of the best SSL models in the literature. 


The original MoCoV3 works by learning invariant representations to color- and spatial-based transformations of an image (e.g. horizontal flip, color jittering \dots). Since we study egocentric videos, we further adapt the model to also learn slowly changing visual representations, following \cite{aubret2022time,pandey2024vision}. For a given input image $x_t$ in a batch, we randomly sample an indirect temporal neighbor $x_{t'}$ within a temporal window $\Delta T$, from the same video recording. The two images capture the same scene from different moments in time, providing a temporally varied view. On these images, we perform a gaze-centered crop of size $N$ (see \cref{sec:dataset}) and apply the standard MoCoV3 augmentation pipeline. Note that it results in two consecutive crop operations per image. The first one extracts the visual input in central vision; the second one randomly crops and resizes a portion of the image within central vision, as part of MoCoV3.

Thereafter, we compute the embeddings of images $q_t=f_q(x_t)$ and $k_{t'}=f_k(x_t')$ using a query feature extractor $f_q$ and a momentum feature extractor $f_k$, both implemented as neural networks. Finally, for a pair $(q_t, k_{t'})$, the query encoder is updated by minimizing the InfoNCE loss \cite{oord2019representationlearningcontrastivepredictive}:
\begin{align}
    \mathcal{L}_{q_t} = - \log \dfrac{ \exp \bigl( \text{sim}(q_t, k_{t'}) / \tau \bigr)}{\sum_{i=0}^K \exp \bigl( \text{sim}(q_t, k_i) / \tau \bigr)}
\end{align}
where \text{sim} denotes cosine similarity, $\tau$ is a temperature hyper-parameter, and $K$ represents the outputs of $f_k$ from the same training batch. Intuitively, the objective increases the similarity between representations of temporally close views ($x_t$ and $x_{t'}$) while enhancing the dissimilarity between all views ($x_t$ and $x_i$).


The momentum encoder gradually updated as an exponential moving average of the query encoder using:
\begin{align}
    \theta_k \leftarrow m \cdot \theta_k + (1 - m) \cdot \theta_q,
\end{align}
where $m$ is a momentum coefficient and $\theta_q$ and $\theta_k$ are the parameters of the query feature extractor and the momentum feature extractor, respectively. We use the \textit{solo-learn} library as an implementation of the model \cite{JMLR:v23:21-1155}.

\subsection{Evaluation}\label{sec:evaluation}
After pre-training our visual encoder, we follow standard SSL transfer learning protocols to evaluate the learned representations. We consider a wide range of downstream tasks that we split into several groups, depending on the semantic ability evaluated by the dataset.

\paragraph{Hard object categorization.} To assess the categorization ability of the models, we consider the ImageNet-1k \cite{russakovsky2015imagenet}, ImageNet100 \cite{tian2020contrastive} and CIFAR100 \cite{krizhevsky2009learning} datasets. We also use two widespread variants of ImageNet-1k with a smaller, balanced training set for the linear probe, namely ImageNet-1k (1\%) and ImageNet-1k (10\%) \cite{chen2020big}.

\paragraph{Easy object categorization.} We also evaluate the models on easier categorization tasks that contain only 10 classes, namely STL10 \cite{coates2011analysis} and CIFAR10 \cite{krizhevsky2009learning}.

\paragraph{Fine-grained object categorization.} Most classes in the two previous groups are for basic-level category recognition (e.g. car, trucks, bananas \dots). Here, we rather assess categorization at the supra-ordinate level (e.g. for cars, differentiating a 2012 Tesla Model S from a 2012 BMW M3 coupe). This requires a model to extract more details about an object. We consider a wide range of supra-ordinate categories: Flowers101 \cite{4756141}, Stanford Cars \cite{6755945}, Oxford Pet \cite{6248092}, FGVC-Aircraft \cite{Maji2013FineGrainedVC}, DTD \cite{6909856}.

\paragraph{Instance object recognition.} We evaluate object instance recognition when exposed in front of different backgrounds with different orientations. We use ToyBox \cite{Wang_2017_ICCV}, COIL100 \cite{nene1996columbia}, Core50 \cite{lomonaco2017core50}. Core50 mostly allows us to assess the robustness of the representation to changing backgrounds, while ToyBox and COIL100 present objects in different positions and orientations. We explain in Appendix how we split the train and test splits.

\paragraph{Scene recognition.} For scene recognition, we focus on Places365-standard \cite{zhou2017places}. This dataset contains $~1.8$ million images from 365 scene categories and is widely used in machine learning. \\

For each dataset, we train a linear classifier on top of the frozen features of the pre-trained encoder for 100 epochs. We apply the standard crop/resize and horizontal flip augmentations during training and report the accuracy on a center crop of validation images. We do not apply center crop on CIFAR10, COIL100 and STL10.

\subsection{Implementation details}

\paragraph{Architecture} 
We employ a ResNet-50 architecture \cite{kaiming2026deep} for all experiments. For MoCoV3 loss, we use a two-layer MLP with a hidden dimension of 4096 to project the output features into a 256-dimensional embedding space. 

\paragraph{Optimization} We train each model for a single epoch on the entire Ego4D dataset. We employ the \textsc{LARS} optimizer \cite{you2017scaling} with a cosine decay learning rate schedule, without restarts, and a warm-up period set to 1\% of the total training steps. The learning rate follows the linear scaling rule proposed by \cite{Goyal2017AccurateLM}, and is given as $lr_{\text{base}} \times B / 256$, where the base learning rate is  $lr_{base}=1.6$ and the total batch size is $B=512$. In addition, we set the weight decay to $1e-6$. We keep the exponential moving average parameter constant at $m = 0.996$ and the temperature of the InfoNCE loss at $\tau = 0.1$. Training is conducted in full precision across 8 GPUs, with a batch size of 64 per GPU.

\section{Experiments}\label{sec:expe}

We aim to assess the impact of focusing on central vision in egocentric SSL. In \Cref{sec:mainres}, we compare training on central vision versus using the whole visual field. Then, we analyze how the size of the field of view of training images impacts learned visual representations (\Cref{sec:field}). Finally, we study the importance of temporal slowness in the context of humans' gaze movements (\Cref{sec:slownessres}) and investigate the role of gaze movements for learning with central vision (\Cref{sec:gazemov}). 

\subsection{Focusing on central vision promotes object-centered representations} \label{sec:mainres}

Here, we compare the impact of learning representations with the whole field of view versus central vision. The whole visual field mimics the training setting used in prior egocentric SSL models \cite{orhan2020self,orhan2024learning,long2024babyview}. In \tableautorefname~\ref{tab:main}, we observe that training with central vision significantly boosts hard category recognition (more than 10 classes). Especially, on the hardest dataset, ImageNet-1k, this leads to an improvement of almost $1.6\%$. Interestingly, we observe an inverse, though weaker, effect when evaluating easy category recognition (less than 10 classes). We suspect that extracting backgrounds that correlate with a given class is often sufficient to perform relatively well on these datasets (e.g. detecting water is enough to classify a boat in cifar-10). We investigate this hypothesis in \Cref{sec:field}. For fine-grained category recognition and instance recognition, training on central vision outperforms training on the full visual field on all evaluated datasets by a substantial margin ($3$-$4\%$ on average). We conclude that egocentric SSL with central vision leads to better object representations.

\begin{table}[]
\caption{Linear probe accuracy on different datasets. For each semantic group of datasets, we show the average recognition accuracy.}
\label{tab:main}
\begin{tabular}{c|c|c}
\toprule
Dataset & Full visual field & \underline{Central vision} \\
\midrule
\multicolumn{3}{c}{Hard category recognition}            \\
\midrule
ImageNet-1k 100\%            & $48.982$      & $\mathbf{50.572}$     \\
ImageNet-1k 10\%            & $35.265$      & $\mathbf{36.277}$      \\
ImageNet-1k 1\%              & $20.114$      & $\mathbf{20.656}$      \\
ImageNet-100              & $70.900$      & $\mathbf{71.463}$      \\
CIFAR100                  & $55.989$      & $\mathbf{56.759}$      \\
Average                   & $46.250$ & $\mathbf{47.145}$\\
\midrule
\multicolumn{3}{c}{Easy category recognition}            \\
\midrule
STL10                     & $\mathbf{71.689}$      & $71.514$      \\
CIFAR10                   & $\mathbf{79.574}$      & $78.654$      \\
Average                   &$\mathbf{75.632}$  & $75.084$   \\
\midrule 
\multicolumn{3}{c}{Fine-grained recognition}            \\
\midrule
DTD                       & $52.176$      & $\mathbf{54.565}$      \\
FGVCAircraft              & $14.418$      & $\mathbf{14.808}$      \\
Flowers102                & $46.163$      & $\mathbf{48.211}$      \\
OxfordIIITPet             & $34.858$      & $\mathbf{47.576}$      \\
StanfordCars              & $21.191$      & $\mathbf{24.111}$      \\
Average                   & $33.761$  & $\mathbf{37.854}$  \\
\midrule 
\multicolumn{3}{c}{Instance recognition}             \\
\midrule
ToyBox                    & $92.593$      & $\mathbf{92.739}$      \\
COIL100                   & $75.563$      & $\mathbf{79.490}$      \\
Core50                    & $25.919$      & $\mathbf{30.437}$      \\
Average                   & $64.691$ & $\mathbf{67.556}$ \\
\bottomrule
\end{tabular}
\end{table}

\subsection{Central vision makes representations less background-sensitive}\label{sec:field}

\begin{figure}
    \centering
    \includegraphics[width=1\linewidth]{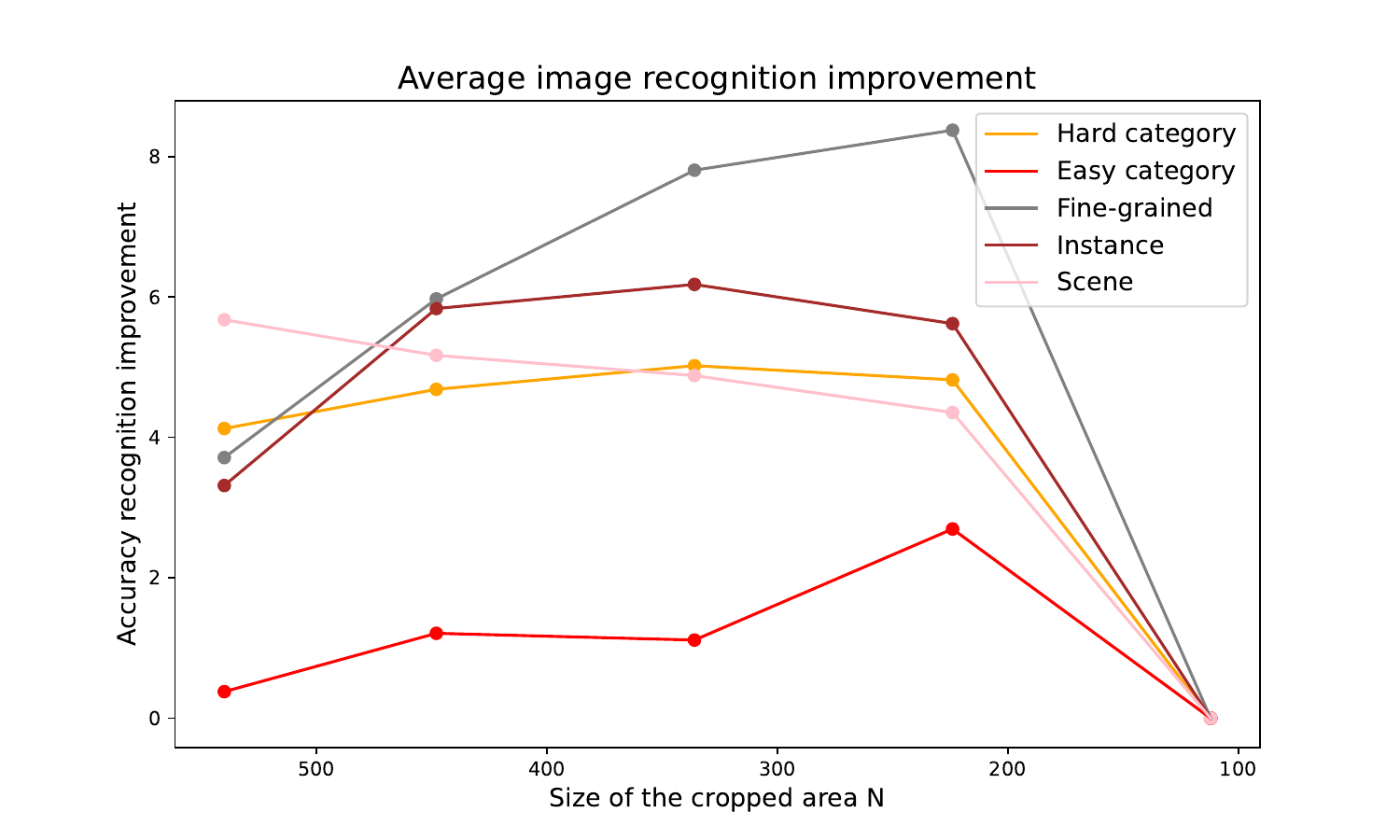}
    \caption{Impact of the gaze-based crop size on different semantic image recognition groups. We compute the average improvement for each semantic group of datasets with respect to $N^2=112 \times 112$. We use a temporal window of $\Delta T=15$ seconds.}
    \label{fig:gsl}
\end{figure}

In this section, we analyze how focusing on central vision affects the learned visual representations. First, we study the impact of the size of the gaze-based crops $N$ on visual learning. In \figureautorefname~\ref{fig:gsl}, we observe a sweet spot in intermediate crop size for all object-centered datasets. This sweet spot is located at $N=336$ for hard category and instance recognition, while $N=224$ seems to be better for easy category and fine-grained recognition. $N=112$ lower-bounds all semantic recognition accuracies, indicating that it probably dismisses too much information about the image. Interestingly, scene recognition accuracy consistently decreases as we make the crops smaller. We conclude that using the whole field of view elicits more scene-based representations, versus object-centered representations for central vision.

Large fields of view tend to display scenes with complex backgrounds and relatively small objects. It may be that extracting background features is easier to satisfy the spatio-temporal invariance objective of the SSL model. To investigate that, we take the ImageNet-9 dataset, a dataset of natural images designed to investigate the background sensitivity of models \cite{xiao2021noise}. We compute the category recognition accuracy of our model with a linear probe trained on ImageNet-1k in all settings (normal image, without background and with different ways to remove the object). We first find that training on central vision also benefits category recognition on normal images for this dataset ($79.83\%$ versus $75.33$). Then, we compute the recognition accuracy when removing the background (Missing background) and when removing the object (Missing object). When removing the object, we average the recognition accuracies of the different ways of removing the foreground object (cf. \cite{xiao2021noise}). To obtain a measure of background and object sensitivity irrespective of the raw performance of the model, we subtract the missing object and missing background accuracies by the recognition accuracy on normal images. 

\begin{figure}
    \centering
    \includegraphics[width=1\linewidth]{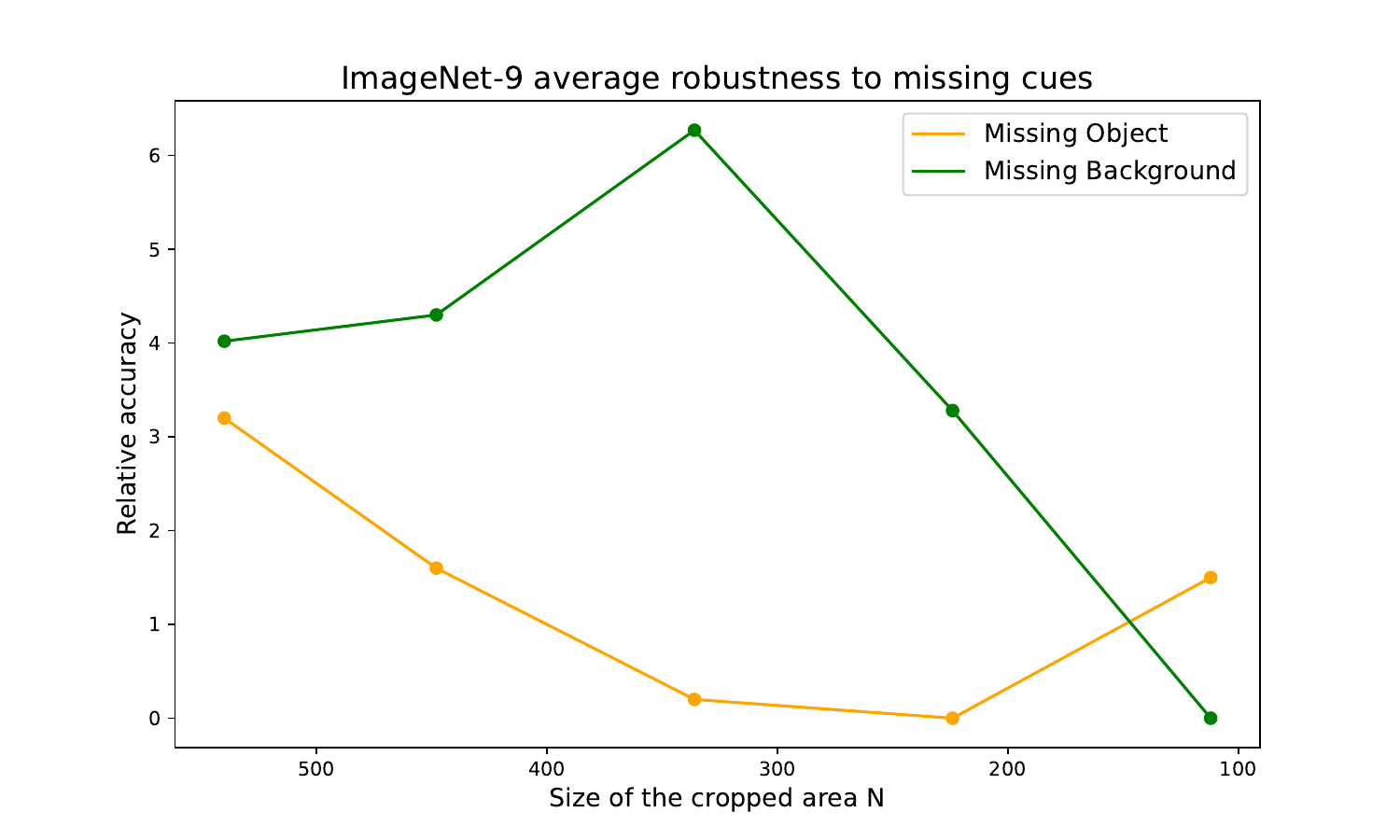}
    \caption{ImageNet-9 recognition sensitivity to missing background or missing foreground object. We show the relative improvement with respect to the worst model for the two settings. The higher, the more relatively robust is the representation to missing backgrounds or missing objects. We use a temporal window $\Delta T=15$ seconds.}
    \label{fig:missingcues}
\end{figure}

In \figureautorefname~\ref{fig:missingcues}, we clearly observe that training on an intermediate size of central vision ($N=336$) allows to rely more on the foreground object (missing background), and less on the background (missing object). We speculate that training on central vision removes a lot of information about the background while often keeping the object intact. For $N=112$, there is an opposite trend, presumably because this is too small to display objects in full. Overall, this suggests that central vision leads to better object-centered representations because it learns to extract less background information.

\subsection{Slowly changing representations boosts SSL with central vision} \label{sec:slownessres}

\begin{figure}
    \centering
    \includegraphics[width=1\linewidth]{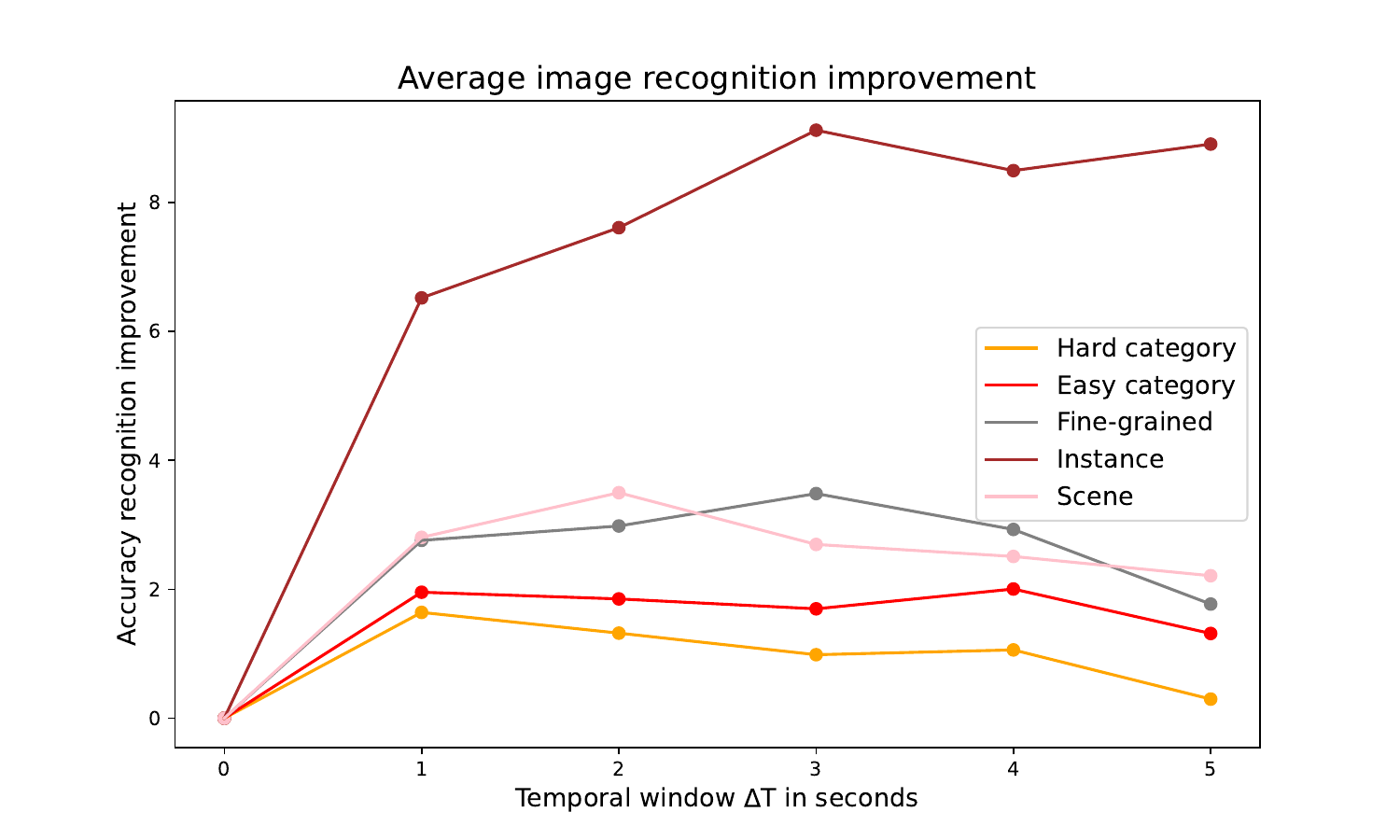}
    \caption{Impact of the temporal window of slowness learning on different semantic image recognition groups. We compute the average improvement for each semantic group of datasets with respect to $\Delta T=0$ second. We use a crop of size $N=224$.}
    \label{fig:slownessresult}
\end{figure}

Previous work on standard videos suggests that learning representations that slowly vary for up to $t=1$ second can be beneficial for visual learning \cite{xu2021rethinking}. However, focusing on gaze-based central vision during egocentric learning provides semantically different temporal dynamics compared to using the whole field of view of, e.g., a movie clip. In \figref{fig:slownessresult}, we present the impact of the level of temporal slowness on visual representations trained with central vision. We observe that temporal slowness is critical for learning representations with respect to all the semantic aspects investigated ($\Delta T=0$ versus $\Delta T > 0$). We provide detailed results in Appendix \tableautorefname~\ref{app:complete}, showing that this improvement is consistent over datasets. However, fine-grained and instance recognition benefit from more slowly changing visual representations (best with $\Delta T=3$ seconds), compared to scene (best at $\Delta T=2)$ seconds) and hard category recognition (best at $\Delta T=1$ seconds). We provide detailed results in Appendix \tableautorefname~\ref{app:complete}, which further show that the best temporal window $\Delta T$ is overall consistent for datasets within a semantic group. We conclude that learning representations that slowly change is a crucial aspect of egocentric SSL with central vision.

\subsection{Humans gaze movements support visual SSL}\label{sec:gazemov}

To investigate the importance of gaze movements to learn about objects, we train our model on a gaze-agnostic visual sequence where we always crop the center of the frame. Note that the resulting visual sequence always varies over time because of the head movements of the camera wearer. In \tableautorefname~\ref{tab:centervscentral}, we observe that training on gaze-based crops consistently outperforms its gaze-agnostic counterparts. This improvement is relatively weak ($< 1\%$) except for instance recognition. However, this average measure is highly influenced by the performance gap on the Core50 dataset (cf. Appendix \tableautorefname~\ref{app:complete}). In sum, using simulated gaze locations of humans (weakly) boosts visual learning.


\begin{table}[]
    \centering
    \begin{tabular}{c|c|c}
        \toprule
        Semantic group & Center crop & \underline{Gaze-based crop} \\
        \midrule
        Hard category rec. & $46.578$ & $\mathbf{46.943}$ \\
        Easy category rec.  & $67.274$  & $\mathbf{68.192}$\\
        Fine-grained rec. & $37.767$ & $\mathbf{38.424}$ \\
        Instance rec. & $62.825$ & $\mathbf{66.997}$ \\
        Scene rec. & $42.691$ & $\mathbf{42.954}$ \\
        \bottomrule
    \end{tabular}
    \caption{Average recognition (rec.) accuracies within semantic groups. We use a crop of size $N=224$ and $\Delta T=15$ seconds}
    \label{tab:centervscentral}
\end{table}



\section{Conclusion}

Inspired by the importance of central vision in humans, we proposed to train SSL models on visual areas corresponding to humans' central vision in head-mounted egocentric video recordings. We simulated humans' gaze locations on the largest-to-date dataset of egocentric videos and extracted the visual areas surrounding the gaze locations. Then, we trained a variant of a mainstream SSL model that learns slowly changing visual representations. 

Our extensive experiments demonstrate that training on central vision generally boosts object representations compared to training on the whole field of view. Especially, this improves hard category, fine-grained and instance object recognition abilities of the models. Our analysis shows that central vision elicits the extraction of more object-related features than background features. In addition, we found that inciting representations to slowly change over time significantly boosts visual learning in all aspects and that gaze locations specifically support the learning process.

 Humans remain far more efficient at learning strong visual representations. For example, the accuracy of the Top-5 linear probe with ImageNet-1k $1\%$ barely goes beyond $40\%$, which is much less than the approximate $90\%$ for humans \cite{russakovsky2015imagenet,orhan2023scaling}. To narrow this gap, a first step would be to simply train bigger models for a longer time \cite{orhan2023scaling}. However, these experiments go beyond the constraints of our computational resources. A second step may be to more realistically mimic biological retinal processing in egocentric SSL models \cite{wang2021use}, thereby inducing a more gradual attenuation of visual information towards the periphery. In practice, it may induce a huge data distribution shift between pre-training images and the evaluation datasets, leaving unclear how to assess such vision models. Yet, our work indicates that central vision is an important component for learning strong object representations. Thus, our work makes a significant step towards narrowing the gap between humans and machines when training on a similar visual experience.
\section*{Aknowledgement}

 This work was funded by the Deutsche Forschungsgemeinschaft (DFG project 5368 ``Abstract REpresentations in Neural Architectures (ARENA)''), as well as the project ``The Adaptive Mind'' funded by the Excellence Program of the Hessian Ministry of Higher Education, Science, Research and Art (HMWK). We gratefully acknowledge support from GENCI–IDRIS (Grant 2022-AD011013678) and Goethe-University (NHR Center NHR@SW) for providing computing and data-processing resources needed for this work. Jochen Triesch was supported by the Johanna Quandt foundation.

{
    \small
    \bibliographystyle{ieeenat_fullname}
    \bibliography{main}
}
\newpage

\appendix
\section{Datasets} \label{app:datasets}
In \tableautorefname~\ref{app:datasets}, we present the datasets and benchmarks used in our experiments. The provided training splits are utilized for training the linear probe, and evaluations are conducted on the test splits. If a test split is unavailable, we use the validation split instead. For Core50, following the approach of \cite{orhan2024learning}, we use 7 backgrounds for training and 5 backgrounds for testing. The task is relatively simple in COIL100, so we train a linear probe on only one image per class.

\begin{table*}[ht]
\centering
\begin{tabular}{lccc}
\toprule
\textbf{Dataset}  & \textbf{Train Split} & \textbf{Test Split} & Citation \\ 
\midrule
\textbf{Hard Category Recognition} & & & \\
ImageNet-1k 100\%  & 1,281,167 (train)  & 50,000 (test) & \cite{russakovsky2015imagenet}\\
ImageNet-1k 10\%  &  128,116 (train) & 50,000 (test) & \cite{chen2020big} \\
ImageNet-1k 1\%   &  12,811 (train) & 50,000 (test) & \cite{chen2020big}\\
ImageNet-100      &  126,689 (train) & 50,000 (test) & \cite{tian2020contrastive}\\
CIFAR100          &  50,000 (train) & 10,000 (test) & \cite{krizhevsky2009learning}\\ 
\midrule
\textbf{Easy Category Recognition} & & & \\
STL10             &  5,000 (train) & 8,000 (test) & \cite{coates2011analysis} \\
CIFAR10           &  50,000 (train) & 10,000 (test) & \cite{krizhevsky2009learning}\\ 
\midrule
\textbf{Fine-Grained Recognition} & & & \\
DTD               & 1,880 (train) & 1,880 (test)  & \cite{6909856}\\
FGVC-Aircraft     & 3,334 (train) & 3,333 (test) & \cite{Maji2013FineGrainedVC}\\
Flowers102        & 1,020 (train) & 6,149 (test)  & \cite{4756141}\\
OxfordIIITPet     & 3,680 (trainval) & 3,669 (test)  & \cite{6248092}\\
StanfordCars      & 8,144 (train) & 8,041 (test) & \cite{6755945}\\ 
\midrule
\textbf{Instance Recognition} & & & \\
ToyBox            &  36,540 (train) & 15,660 (test) & \cite{Wang_2017_ICCV}\\
COIL100           &  100 (train) & 7,100 (test) & \cite{nene1996columbia}\\
Core50            &  90,000 (train) & 75,000(test)  & \cite{lomonaco2017core50}\\ 
\midrule
\textbf{Scene Recognition} & & & \\
Places365         & 1,803,460 (train) & 36,500 (test)  & \cite{zhou2017places}\\ 
\bottomrule
\end{tabular}
\caption{Overview of datasets, including the number of datapoints in each split, the splits used for training a linear probe, and the splits used for evaluation.}
\label{tab:dataset_splits}
\end{table*}

\section{Gaze Center Location}
\begin{figure}[h]
     \centering
     \includegraphics[width=1\linewidth]{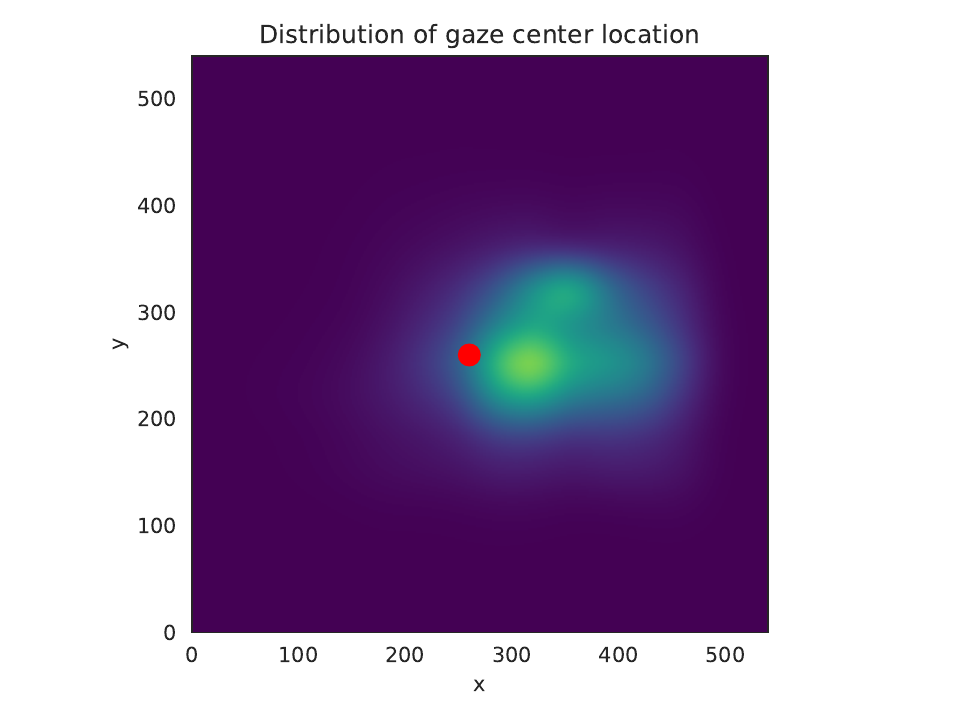}
     \caption{Distribution of the gaze center location over the Ego4D dataset. The red dot symbolizes the center of the frame.}
     \label{fig:gaze_heatmap}
 \end{figure}

In \figref{fig:gaze_heatmap}, we present the distribution of gaze centers across the entire training dataset of Ego4D. Although the gaze prediction model is biased toward the center, there is a shift toward the right side of the image. This suggests that a gaze-centered crop captures subtly different aspects of an image compared to a simple center crop, which can benefit object recognition.

\section{Complete results data} \label{app:complete}
We show in \tableautorefname~\ref{tab:completetab} and \tableautorefname~\ref{tab:completetab2} the detailed results of Figure~2 and Table~2, respectively. Our results are overall consistent in our semantic groups. 

\begin{table*}[] 
\centering
\label{tab:slowres}
\begin{tabular}{cccccccc}
\toprule
   Dataset                & $t=0$           & $t=1$          & $t=2$          & $t=3$          & $t=4$         & $t=5$         \\
   \midrule
\multicolumn{7}{c}{\textbf{Hard category recognition}}   \\
ImageNet-1k 100\%         & $48.642$      & $\mathbf{50.178}    $ & $49.600     $ & $49.578     $ & $49.122    $ & $48.904$     \\
ImageNet-1k 10 \%         & $34.533$      & $\mathbf{35.985}    $ & $35.623     $ & $35.337     $ & $35.117    $ & $34.771$     \\
ImageNet-1k 1\%           & $18.516$      & $\mathbf{20.374}    $ & $20.286     $ & $20.250     $ & $19.738    $ & $19.846$     \\
ImageNet-100              & $69.080$      & $\mathbf{71.260}    $ & $70.940     $ & $70.340     $ & $70.900    $ & $70.860$     \\
CIFAR100                  & $59.018$      & $\mathbf{60.198}    $ & $59.938     $ & $59.208     $ & $60.208    $ & $56.889$     \\
Average                   & $45.958$  & $\mathbf{47.599}$ & $47.277 $ & $46.943$ & $47.017 $ & $46.254$  \\
\midrule
\multicolumn{7}{c}{\textbf{Easy category recognition}}           \\
STL10                     & $66.567$      & $70.215    $ & $70.927     $ & $\mathbf{71.189}     $ & $70.715    $ & $70.915$     \\
CIFAR10                   & $80.554$      & $\mathbf{80.814}    $ & $79.894     $ & $79.324     $ & $80.414    $ & $78.834$     \\
Average                   & $73.560$ & $75.514 $ & $75.410   $ & $75.257  $ & $\mathbf{75.564}$ & $74.874$  \\
\midrule
\multicolumn{7}{c}{\textbf{Fine-grained recognition}}          \\
DTD                       & $46.391$      & $54.777    $ & $55.520     $ & $\mathbf{57.059}     $ & $56.582    $ & $55.414$     \\
FGVCAircraft              & $14.748$      & $14.209    $ & $14.928     $ & $\mathbf{15.767}     $ & $14.838    $ & $13.639$     \\
Flowers102                & $46.504$      & $48.894    $ & $47.740     $ & $49.008     $ & $\mathbf{49.041}    $ & $47.317$     \\
OxfordIIITPet             & $46.351$      & $48.203    $ & $\mathbf{48.339}     $ & $47.032     $ & $45.861    $ & $44.499$     \\
StanfordCars              & $20.706$      & $22.409    $ & $23.080     $ & $\mathbf{23.254}     $ & $23.018    $ & $22.682$     \\
Average                   & $34.940$  & $37.698$ & $37.921  $ & $\mathbf{38.424} $ & $37.868$ & $36.710235$  \\
\midrule
\multicolumn{7}{c}{\textbf{Instance recognition}}       \\
ToyBox                    & $86.986$      & $90.990    $ & $92.286     $ & $\mathbf{92.612}     $ & $92.561    $ & $92.516$     \\
COIL100                   & $69.665$      & $78.364    $ & $80.054     $ & $80.124     $ & $\mathbf{82.430}    $ & $79.786$     \\
Core50                    & $16.979$      & $23.835    $ & $24.116     $ & $\mathbf{28.256}     $ & $24.120    $ & $28.043$    \\
Average                   & $57.876$ & $64.396 $ & $65.485$ & $\mathbf{66.997}$ & $66.370$ & $66.781$ \\
\midrule
\multicolumn{7}{c}{\textbf{Scene recognition}}    \\
Places365      &  $40.259$ & $43.064$ & $\mathbf{43.757}$ & $42.954$ & $42.768$ & $42.469$            \\
\bottomrule
\end{tabular}
\caption{Detailed results of Figure~2. Top-1 accuracy on different datasets when training from different time windows.}
\label{tab:completetab}
\end{table*}

\begin{table*}[]
\centering
\begin{tabular}{ccc}
\toprule
Dataset & Center crop & \underline{Gaze-based crop} \\
\midrule
\multicolumn{3}{c}{\textbf{Hard category recognition}} \\
ImageNet-1k 100\%     & 49.094 & \textbf{49.578} \\
Imagenet-1k 10 \% & 34.891 & \textbf{35.337} \\
ImageNet-1k 1\%   & 19.866 & \textbf{20.250} \\
ImageNet-100   & \textbf{70.460} & 70.340 \\
CIFAR100       & 58.578 & \textbf{59.208} \\
Average        & 46.578 & \textbf{46.943} \\
\midrule
\multicolumn{3}{c}{\textbf{Easy category recognition}} \\
STL10          & 70.590 & \textbf{71.189} \\
CIFAR10        & 79.134 & \textbf{79.324} \\
Average        & 74.862 & \textbf{75.257} \\
\midrule
\multicolumn{3}{c}{\textbf{Fine-grained recognition}} \\
DTD            & 55.414 & \textbf{57.059} \\
FGVCAircraft   & 14.658 & \textbf{15.767} \\
Flowers102     & 48.537 & \textbf{49.008} \\
OxfordIIITPet  & \textbf{47.086} & 47.032 \\
StanfordCars   & 23.142 & \textbf{23.254} \\
Average        & 37.767 & \textbf{38.424} \\
\midrule
\multicolumn{3}{c}{\textbf{Instance recognition}} \\
ToyBox         & 91.769 & \textbf{92.612} \\
COIL100        & 79.420 & \textbf{80.124} \\
Core50         & 17.287 & \textbf{28.256} \\
Average        & 62.825 & \textbf{66.997} \\
\midrule
\multicolumn{3}{c}{\textbf{Scene recognition}} \\
Places375      & 42.691 & \textbf{42.954} \\
\bottomrule
\end{tabular}
\caption{Detailed results of Table~2. Top-1 linear accuracy when pre-training with center versus gaze-based visual cropping.}
\label{tab:completetab2}
\end{table*}


\end{document}